\pdfoutput=1

\documentclass[11pt]{article}

\usepackage[final]{acl}

\usepackage{times}
\usepackage{latexsym}

\usepackage[T1]{fontenc}

\usepackage[utf8]{inputenc}

\usepackage{microtype}

\usepackage{inconsolata}

\usepackage{graphicx}
\usepackage{subcaption}
\usepackage{stfloats}

\title{Thinking Before You Speak: A Proactive Test-time Scaling Approach}

\author{Cong Liu, Wenchang Chai$^\dagger$, Hejun Wu, Yan Pan, Pengxu Wei, Liang Lin \\
  Sun Yat-sen University, $^\dagger$Hong Kong Polytechnic University \\
  \texttt{liucong3@mail.sysu.edu.cn, wenchang.chai@connect.polyu.hk,} \\ 
  \texttt{wuhejun.sysu.edu.cn, panyan5@mail.sysu.edu.cn, } \\ 
  \texttt{weipx3@mail.sysu.edu.cn, linliang@ieee.org} \\}

\begin{document}
\maketitle


\begin{abstract}

Large Language Models (LLMs) often exhibit deficiencies with complex reasoning tasks, such as maths, which we attribute to the discrepancy between human reasoning patterns and those presented in the LLMs' training data. When dealing with complex problems, humans tend to think carefully before expressing solutions. However, they often do not articulate their inner thoughts, including their intentions and chosen methodologies. Consequently, critical insights essential for bridging reasoning steps may be absent in training data collected from human sources. To bridge this gap, we proposes inserting \emph{insight}s between consecutive reasoning steps, which review the status and initiate the next reasoning steps. Unlike prior prompting strategies that rely on a single or a workflow of static prompts to facilitate reasoning, \emph{insight}s are \emph{proactively} generated  to guide reasoning processes. We implement our idea as a reasoning framework, named \emph{Thinking Before You Speak} (TBYS), and design a pipeline for automatically collecting and filtering in-context examples for the generation of \emph{insight}s, which alleviates human labeling efforts and fine-tuning overheads. Experiments on challenging mathematical datasets verify the effectiveness of TBYS. Project website: https://gitee.com/jswrt/TBYS
\end{abstract}

\begin{figure}[!htbp]
  \centering
  \includegraphics[trim=41 117 610 86,clip,width=0.45\textwidth]{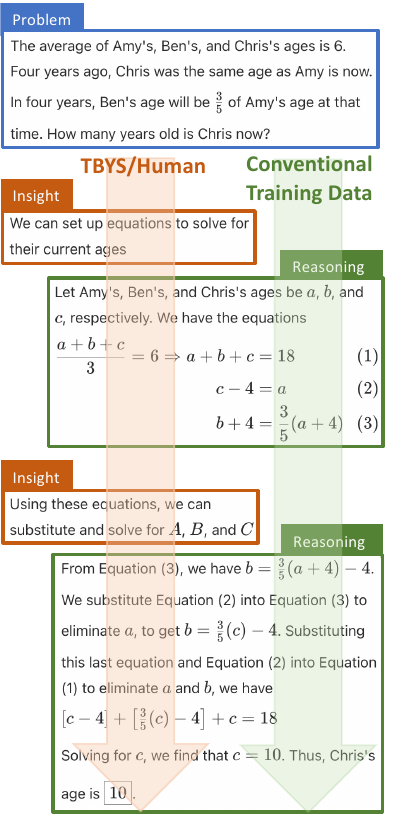}
  \caption{A simplified example to compare the reasoning trace of human and TBYS against one from conventional training data. Humans and TBYS excel with a flow of insight-driven reasoning that is more comprehensible. On the other hand, the training set example adds to the difficulty of learning, as it is not always straightforward to re-engineer the connection between consecutive steps behind the succinct reasoning logic. TBYS proactively fill reasoning gaps with \emph{insight}s representing intention, explanation, or justification, etc. }
  \label{fig:math-example}
\end{figure}

\section{Introduction}

OpenAI’s O1 \cite{o1} demonstrates the potential of leveraging long chains of thought (CoT) \cite{wei2022chain} to enhance the reasoning capabilities of large language models (LLMs). Through its generated reasoning, O1 exhibits advanced cognitive skills, such as problem decomposition, error identification, and correction -- processes that continuously guide thinking toward accurate solutions. Inspired by this, various test-time scaling \cite{scaling-testtime, survey2025} approaches were proposed, such as using prompts like “Wait,” \cite{s1} to stimulate self-correction, “Wait, using Python” to encourage coding \cite{START}, or fixed workflows of prompts to structure inferences \cite{MetaGPT}. However, these methods suffer from task and LLM sensitivity. For instance, certain agentic workflows (e.g., MetaGPT \cite{MetaGPT}) may improve coding tasks but not Q\&A performance. Similarly, LLMs exhibit sensitivity to prompt design, including style and example ordering \cite{prosa}. As a result, they are most effective when paired with reinforcement learning techniques (e.g., rejection sampling) to filter suboptimal cases, but are ill-suited for direct application to scale reasoning at test time.

This paper introduces a novel prompting paradigm called \textbf{proactive prompting}, where an LLM proactively generates prompts to steer its own reasoning steps, rather than passively reacting to predefined prompting patterns. This approach demonstrates particular advantages in complex reasoning tasks, such as advanced mathematics problems, where the proactive generation of ``inner thoughts'' (critical for guiding reasoning) is often absent from final reasoning outputs in conventional training data.

To validate this paradigm, we develop a reasoning framework named \emph{Thinking Before You Speak} (\textbf{TBYS}), which iteratively inserts a proactive prompt -- termed the \emph{insight} -- before each reasoning step to explicitly define the status and the goal of that step. Figure~\ref{fig:math-example} contrasts a TBYS reasoning process with that in conventional training data (with which LLMs are trained). TBYS mirrors human inner-thinking patterns, producing more explainable reasoning traces that facilitate LLM learning and offering greater educational values for human readers.

In the remainder of this paper, we detail the TBYS reasoning framework in Section~\ref{section:tbys}. Since TBYS relies on iteratively generating insights to guide reasoning, the quality of these generated insights is critical to its accuracy. To address this, we employ in-context learning with examples retrieved from a library of insight exemplars. Section~\ref{section:insight} describes our pipeline for automatically collecting, filtering, and selecting example \emph{insight}s for this library. Section~\ref{section:related-work} briefly reviews prior related work. Finally, Section~\ref{section:experiments} evaluates TBYS against strong baselines on challenging datasets, demonstrating significant performance improvements and better accuracy-overhead trade-offs. We further conduct ablation studies to validate the contributions of key components. 

\begin{figure}[h]
  \centering
  \includegraphics[trim=45 191 532 86,clip,width=0.45\textwidth]{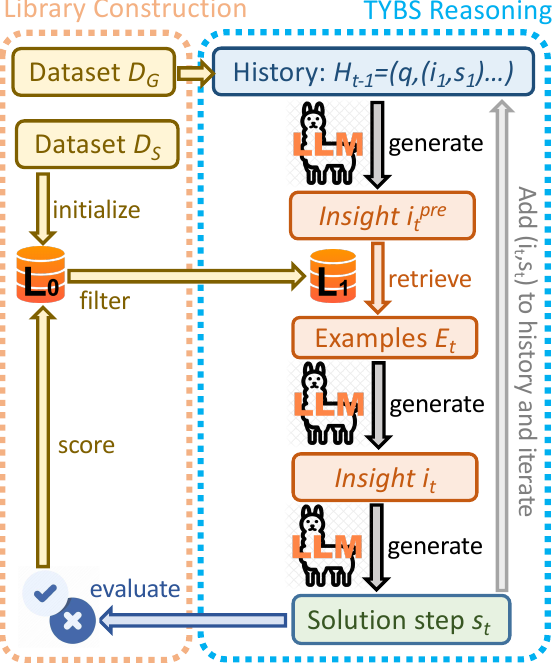}
  \caption{The TBYS reasoning framework (Section~\ref{section:tbys}) and \emph{insight} library construction (Section~\ref{section:insight}).}
  \label{fig:tbys}
\end{figure}

\section{The TBYS Reasoning Framework} \label{section:tbys}

TBYS utilizes a library $L$ of high-quality \emph{insight}s. The automatic construction of this library is detailed in Section \ref{section:insight}. During inference, examples are retrieved from $L$ using some off-the-shelf embedding model for in-context learning. We also manually define three seed examples $S$, each containing a question and the complete reasoning steps for the question with the associated \emph{insight}s.

As shown in Figure \ref{fig:tbys}, TBYS employs a multi-round reasoning approach. Each round $t$ consists of three steps: (1) \emph{Insight Generation}: A preliminary \emph{insight} $i_t^{pre}$ is generated based on the current reasoning history $H_{t-1} = (q, (i_1,s_1),(i_2,s_2),\cdots,(i_{t-1},s_{t-1}))$, where $q$ is the question, and $i_j, s_j$ denote the \emph{insight} and solution step in round $j$, respectively. (2) \emph{Example Retrieval}: Each \emph{insight} is defined by its two components: \emph{situation} (summarizing the current reasoning status) and \emph{goal} (stating the intention for solution step $s_t$). The situation of $i_t^{pre}$ is used to retrieve $k_E=8$ examples $E_t$ from library $L$. Using these $k_E$ high-quality \emph{insight}s as in-context examples, a refined \emph{insight} $i_t$ is generated. (3) \emph{Solution Step Generation}: The solution step $s_t$ is generated using $H_{t-1}$ and $i_t$, then appended to $H_{t-1}$ to form $H_t$. To signal the end of reasoning, $s_t$ includes a field indicating whether a confident answer to $q$ has been reached.

\begin{figure*}[b]
  \centering
  \begin{minipage}{0.48\textwidth}
    \centering
    \includegraphics[width=1\textwidth]{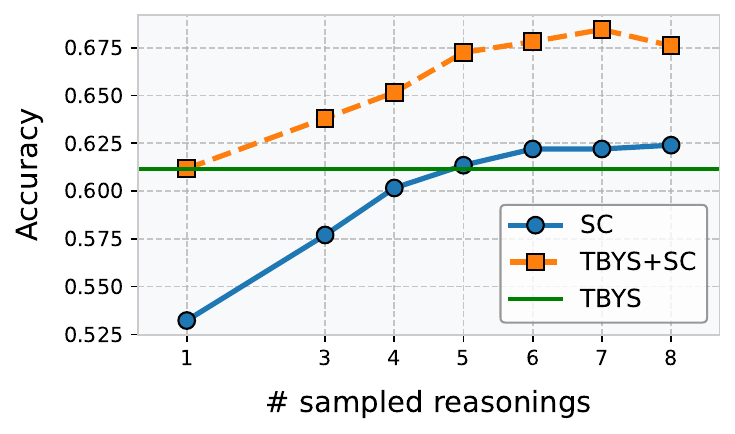}
    \caption{Performance comparison on MATH-500} \label{fig:cmp_math500}
  \end{minipage}
  \hfill
  \begin{minipage}{0.48\textwidth}
    \centering
    \includegraphics[width=1\textwidth]{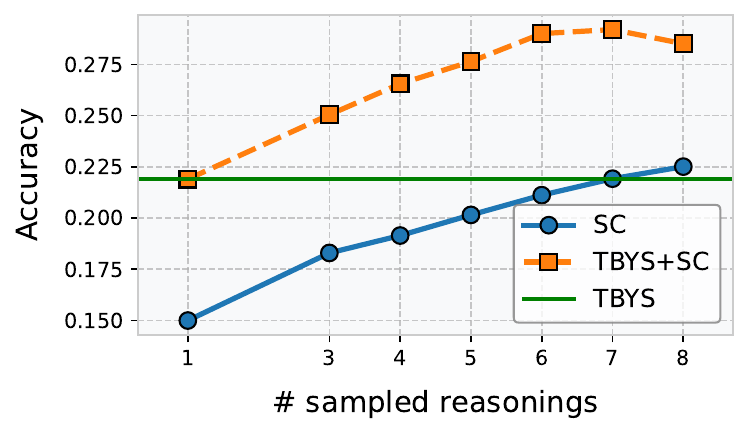} 
    \caption{Performance comparison on AIME} \label{fig:cmp_aime}
  \end{minipage}
\end{figure*}

\section{Construction of the \emph{Insight} Library} \label{section:insight}

As shown in Figure~\ref{fig:tbys}, we build the library of \emph{insight}s in two stages: initialization and filtering. 

\emph{Initialization}: We use manually curated seed examples $S$ and a dataset $D_S$ containing questions and their chain-of-thought solutions. First, an LLM is prompted to split each solution in $D_S$ into 1–3 steps. The LLM is then prompted again to generate an \emph{insight}  $i_t$ for each solution step $s_t$, consisting of a \emph{situation}, which should represents the reasoning status up to that step, and a \emph{goal}, which should offers a purpose and a guideline to stimulate the LLM to reproduce solution step $s_t$. All \emph{insight}s and divided solution steps are collected into an initial library $L_0$.

\emph{Filtering}: To identify high-quality \emph{insight}s, we use a dataset $D_G$ (containing questions and ground-truth answers) and a scoring mechanism: (1) For each \emph{insight} $i_i \in L_0$, maintain counters $r_i$ (correct uses) and $w_i$ (wrong uses). (2) Evaluate $L_0$ by running \emph{TBYS} on each question $q \in D_G$. For each reasoning step for $q$, retrieve $k_F = 25$ examples from $L_0$ and randomly select one as a 1-shot example. If the reasoning yields a correct answer, increment $r_i$ for each $i_i$ used; otherwise, increment $w_i$. (3) Rank \emph{insight}s in $L_0$ by the score $\frac{r_i}{r_i + w_i} \log(r_i + w_i)$, which balances accuracy and usage coverage. Select the top-$k_L$ examples to form $L_1$. The insight library can be progressively improve through multiple iterations. In each iteration, the initial library $L_0$ is updated to include the filtered library $L_1$ and the newly generated insights from dataset $D_G$, which is produced during the filtering of $L_1$.

In our experiments, the MATH-500 dataset \cite{MATH-500} serves as \(D_S\) and the test set, e.g., MATH-500 or AIME \cite{aime}, serves as \(D_G\) in a test-time adaptation \cite{Test-Time-Adaptation} manner, with $k_L$ as a variable parameter.

\section{Related Work} \label{section:related-work}

Extensive research has investigated prompt designs to improve LLM reasoning, including \emph{Chain-of-Thought} \cite{wei2022chain}, \emph{Least-to-Most} \cite{zhou2023leasttomost}, \emph{Self-Consistency} \cite{wang2023selfconsistency}, and \emph{Tree-of-Thoughts} \cite{prob-tree-of-thought}. Methods to enhance task-specific performance include question rephrasing, subtask decomposition, verification, and symbolic grounding \cite{faithful-chain-of-thought, symbolic-chain-of-thought, plan-and-solve, star, CoK}; factuality and faithfulness checking for reasoning chains \cite{CoK}; and separating knowledge retrieval from reasoning \cite{disentangle}. 

Iterative prompting techniques rely on predefined, hardcoded actions to guide reasoning, such as \emph{Self-Refine} \cite{self_refine}, \emph{IRCoT} \cite{IRCoT}, \emph{iCAP} \cite{iCAP}, \emph{MetaGPT} \cite{MetaGPT}, and \emph{Chain of Ideas} \cite{CoI}.

Memory-based methods include \emph{Buffer of Thoughts} \cite{buffer-of-thoughts}, which distills high-level guidelines from previously solved tasks and stores them in a buffer for future reuse. Skill-based CoT \cite{metacognitive-math} predicts skill-based labels for the questions. \cite{induction-augmented-generation} identifies key concepts in questions and uses inductive prompting templates to extract related concepts. 

\emph{rStar} \cite{rStar} employs a self-play mutual reasoning approach, augmented by Monte Carlo Tree Search (MCTS) with a set of five reasoning-inducing prompts, to enhance reasoning. 

Finetuning-based methods, such as \emph{STaR} \cite{star}, \emph{ReST-MCTS} \cite{ReST-MCTS}, and \emph{AFlow} \cite{aflow}, demonstrate that iterative training on reasoning histories and task-specific workflows of correct answers enables models to tackle increasingly complex problems.

\section{Experiments} \label{section:experiments}

\subsection{Experiment settings}

We conducted experiments on two challenging mathematical datasets, \emph{AIME} \cite{aime} and MATH-500 \cite{MATH-500}. We compare TBYS against a simple yet very strong baseline: 8-shot \emph{In-context Learning} \cite{Ordered-Prompts} with \emph{Self-Consistency} \cite{wang2023selfconsistency}. 

For the experiments, use utilize the LLM \emph{Qwen2.5-7B-Instruct} \cite{qwen2} via the LLM API provided by Siliconflow \cite{siliconflow}, with the following configurations: max\_tokens=1024, temperature=0.2, top\_k=40, top\_p=0.7, and n=1. The \emph{bge-large-en-v1.5} embedding model is employed for \emph{insight} retrieval. Results are reported as the average across 8 experimental runs.

Since coding benefits mathematical problems \cite{PoT}, when Python code blocks are detected in the LLMs' responses, we invoke a customized sandboxed Python interpreter and append the output to the code block.

\subsection{Comparison}

When compared with \emph{Self-Consistency} (SC), TBYS demonstrates comparable performance to SC using 5 reasoning samples (SC@5) on MATH-500 (Figure~\ref{fig:cmp_math500}) and SC@7 on AIME (Figure~\ref{fig:cmp_aime}). The results further indicate that TBYS integrates effectively with SC: TBYS+SC yields over 5\% absolute gains in accuracy on MATH-500 and 7.5\% on AIME.

\subsection{Overhead Analysis}

\begin{table}[h]
\centering
\small
\caption{Cost comparison to SC under similar accuracy.}
\label{tab:overhead}
\begin{tabular}{|r|c|c|c|c|}
\hline
\textbf{\scriptsize{MATH-500}} & \textbf{Acc.}  & \textbf{Time} & \textbf{Prompt} & \textbf{Completion} \\
\hline
TBYS & 0.61    & \underline{52.82}    & 18163.80 & \underline{999.57}    \\
SC@5 & 0.61 & 102.56  & \underline{13334.62} & 2217.30   \\
\hline
\hline
\textbf{AIME} & \textbf{Acc.}  & \textbf{Time} & \textbf{Prompt} & \textbf{Completion} \\
\hline
TBYS & 0.22    & \underline{78.15}  & \underline{20686.23} & \underline{1559.60}   \\
SC@7 & 0.22   & 322.79  & 25,242.54 & 7,102.49   \\
\hline
\end{tabular}
\end{table}

We compare the overhead of TBYS with SC@5 on MATH-500 and with SC@7 on AIME, where the methods achieve comparable accuracies. The metrics analyzed include wall-time, number of prompt tokens, and completion tokens. As shown in Table~\ref{tab:overhead}, under similar accuracies, TBYS reduces wall-time and the number of completion tokens by approximately half on MATH-500 and one-third on AIME. While TBYS uses 46\% more prompt tokens on MATH-500, these can be cached and are typically much cheaper and faster to predict than completion tokens. Since completion token counts typically dominates runtime, our results show that completion token counts are consistently proportional to our runtime measurements across methods.

\subsection{Ablation Study}

\begin{table}[h]
\centering
\small
\caption{Ablation Study}
\label{tab:ablation}
\begin{tabular}{|l|c|c|}
\hline
 & \textbf{MATH-500} & \textbf{AIME} \\
\hline
TBYS & 61.17\% & 21.90\% \\
~~~- Library Construction & 58.90\% & 19.51\% \\
~~~- Coding & 57.00\% & 18.11\% \\
\hline
8-shot & 53.23\% & 14.99\% \\
\hline
\end{tabular}
\end{table}

We conducted ablation experiments by using the raw insight library \(L_0\) as \(L_1\) (without filtering, as described in Section \ref{section:insight}). Accuracy declines were observed in both datasets. Notably, we only performed one round of insight filtering (i.e., using \(L = L_1\)), and additional filtering rounds are expected to further improve accuracy. Table \ref{tab:ablation} also demonstrates that coding contributes half of the accuracy gain compared to simple 8-shot prompting.

\subsection{Impact of Library Size}

\begin{figure}[h]
  \centering
  \begin{minipage}{0.48\textwidth}
    \centering
    \includegraphics[width=\textwidth]{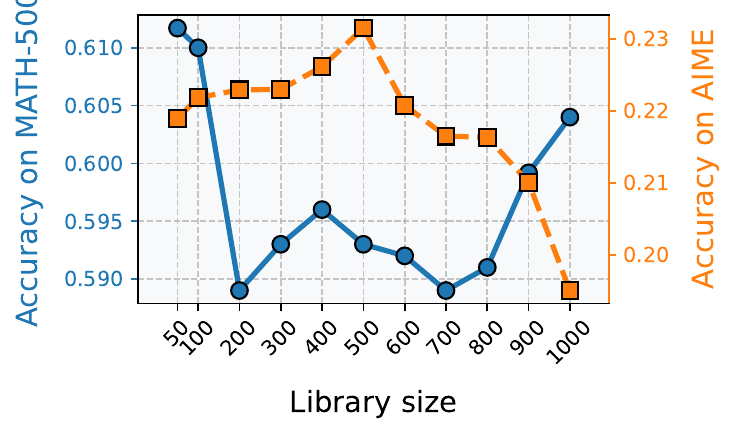}
    \caption{Impact of \emph{insight} library size} \label{fig:size_lib}
  \end{minipage}
\end{figure}

In Section \ref{section:insight}, we sorted the \emph{insight} library $L_0$ and selected the top-\(k_L\) insights to form \(L_1\). Figure \ref{fig:size_lib} shows that on MATH-500, TBYS achieves peak accuracy with an insight library size of 50.  On AIME, the optimal size is 500. Here, performance initially improves as library size decreases due to the filtering of lower-quality insights. However, as library size continue to decreases, excessively small libraries size reduces diversity in problem types and harms performance.




\section{Additional Comparison Experiments}

We compare with Skill-based CoT \cite{metacognitive-math}, a prompt-guided interaction procedure that enables LLMs to assign skill labels to math questions and perform ICL with label-specific examplars. We also conducted experiments using a $k$-wait approach, where we append “Wait, ” after the model completion and let the LLM to continue its generation for k times. Below are the comparison results.

\begin{table}[h]
\caption{Comparison to Skill-based CoT and $k$-wait.} \label{cmp2.fig}
    \centering
    \begin{tabular}{lc}
        \hline
        Method           & Acc.   \\
        \hline
        k-shot CoT       & 54.30\% \\
        Skill-based CoT  & 60.52\% \\
        k-wait (k=1)     & 55.00\% \\
        k-wait (k=2)     & 56.60\% \\
        k-wait (k=3)     & 54.20\% \\
        TBYS (Ours)      & 61.99\% \\
        \hline
    \end{tabular}
\end{table}

Results in Figure~\ref{cmp2.fig} shown that TBYS is slight better than Skill-based CoT, which is task-specific, and is much better than $k$-wait.

\section{Qualitative Analysis of Insight Quality}

We provide qualitative analysis of the insights using two selected examples. These problems are relatively simple, but where $k$-shot reasoning fails. We use these examples to illustrate how TBYS's insights effectively steer its multi-step reasoning processes.

The first example in Figure~\ref{fig:example1} asks to convert $\frac{21}{2^2 \cdot 5^7}$ to a terminating decimal. TBYS solves the problem in two steps, with the goals of the insights being ``ensure the denominator is a power of 10'' and ``Simplify the numerator and express the fraction as a terminating decimal''.

The second example in Figure~\ref{fig:example2} asks to solve the question: $\sqrt{x + \sqrt{3x + 6}} + \sqrt{x - \sqrt{3x + 6}} = 6$. TBYS solves the problem in two steps, with the, with the goals of the insights being ``simplify the equation by squaring both sides to remove the square roots'' and ``find the value of $x$ by substitution''.

Both examples demonstrate that TBYS generates suitable insights for their respective problems.

\section{Conclusion and Future Work}

This paper introduces a novel proactive prompting paradigm, instantiates it with the simple TBYS reasoning framework, and verifies the effectiveness of TBYS on challenging advanced mathematics reasoning tasks.

Promising directions for future improvement include: Automated search for optimal \emph{insight}s \cite{llm-optim}; integration of long-term memory mechanisms \cite{ChemAgent, rag}; enhancement of programming capabilities \cite{PoT}; enforcement of structured inference processes \cite{StructRAG, prob-tree-of-thought}.

\section*{Limitations}

Our method incurs higher computational overhead compared to direct prompting, a common drawback among advanced prompting techniques that involve scaling test-time inference.

Due to time and financial constraints (our current experiments take about 50 days with single-threaded API calls), we only evaluated the proposed method on two math-domain datasets using a single LLM.

\section*{Ethical Statement}

This work fully adheres to the ACL Ethics Policy. To the best of our knowledge, no ethical issues are associated with this research.

\bibliography{ref}

\appendix

\section{Figures for Qualitative Analysis of Insight Quality}

\begin{figure*}[h]
  \centering
    \includegraphics[trim=30 500 25 30, clip, width=0.75\textwidth]{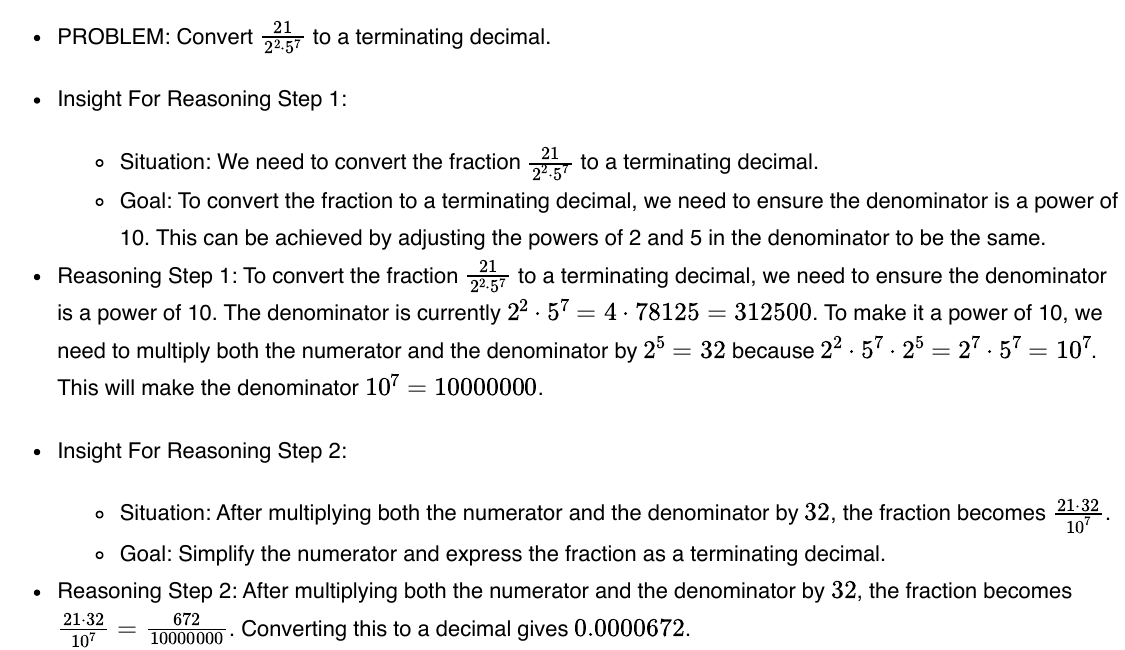}
    \caption{Example 1 for Qualitative Analysis of Insight Quality} \label{fig:example1}
\end{figure*}

\begin{figure*}[h]
  \centering
    \includegraphics[trim=30 240 25 30, clip, width=0.94\textwidth]{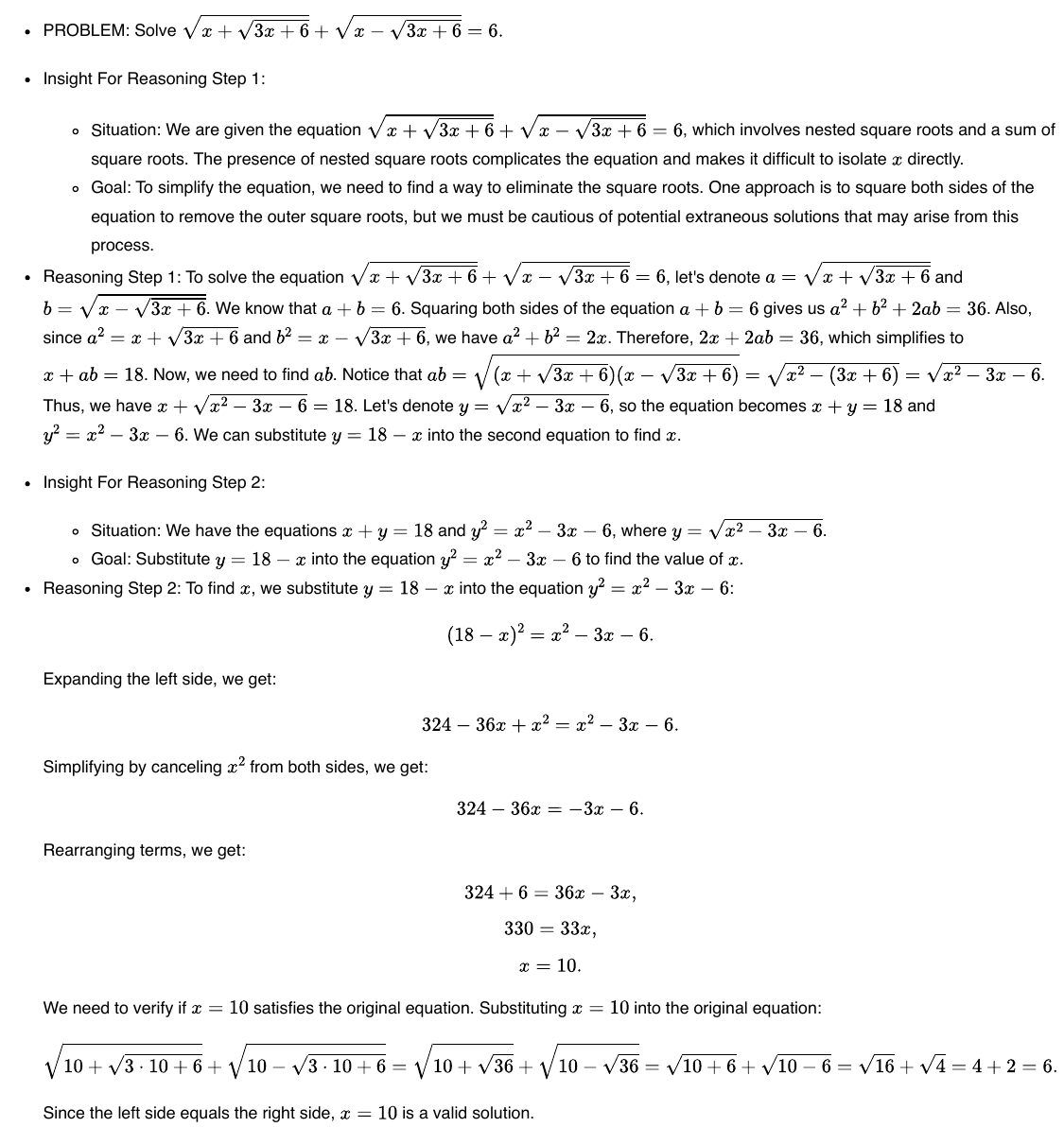}
    \caption{Example 2 for Qualitative Analysis of Insight Quality} \label{fig:example2}
\end{figure*}

\end{document}